# Automatic Part-of-Speech Tagging of Arabic-English Dictionary Senses through WordNet

Diaa M. Fayed[*1], Aly A. Fahmy[*2], Mohsen A. Rashwan[**3], Wafaa K. Fayed[***4]

[*]*Computer Science, Faculty of Computers and Information, Giza, Egypt*

[1]diaa.fayed@outlook.com

[2]a.fahmy@fci-cu.edu.eg

[**]*EECE, Faculty of Engineering, Giza, Egypt*

[3]*mrashwan@ieee.org*

[***]*Arabic Language and Literatures, Faculty of Arts, Giza, Egypt*

[4]*wafkamel@link.net*

***Abstract***— This paper proposed an algorithm for part-of-speech (POS) tagging senses of a bilingual dictionary. The algorithm is applied on the Al-Mawrid Arabic-English dictionary. The tagging task is accomplished by transferring the POS tags of the English translation equivalences (TEs) to the dictionary senses after dis-ambiguities process. The English POS tags of senses are acquired from the Princeton WordNet. POS tagging of bilingual dictionary senses is prerequisite to link a bilingual dictionary to WordNet and/or standardizing that dictionary into WordNet-LMF format where the synset (set of synonyms), not word, is the basic brick. The registered accuracy is high though the cost is little. Building NLP/HLT tools needs linguistic experts, large investments, and long time. For statistical approach, we need large annotated corpora and for rule-based approach, we need large lexicon that contains rich linguistic and world knowledge. That motivates the appearance of what are called resource-light approaches to develop natural language processing (NLP) tools for poor-resource languages.

## 1 INTRODUCTION

Vast researches and investments were made for Latin languages specially English in the field of natural language processing (NLP) and human language technology (HLT). The results of these are much amount of resources and tools such are lexicons, thesauruses, annotated corpora, morphological analyzers, syntactic parsers, etc. On the other hand, other languages, such as Arabic, are poor of those resources and tools. Some researches such as parallel text processing try to benefit from resources and tools that are built for Latin languages to build resources and tools for other poor-resources languages [1-11]. Yarowsky and Ngai [12] stated that we can overcome on resource shortage problem of some languages by leveraging the annotated data and tools for resource-rich languages (such English, French and Japanese).

Feldman [7] summarized resource-light approaches to NLP tasks as unsupervised or minimally supervised approaches and cross-language knowledge induction. Instances of the former approach are unsupervised POS tagging and minimally supervised morphology learning. Instances of the latter approach are cross-language knowledge transfer using parallel texts, bilingual lexicon acquisition, and cross-language knowledge transfer without parallel corpora.

Annotated language sources such as corpora and dictionaries are required in both HLT and NLP. The annotations are any information that augmented to text so as to make computer to either understand the text or used in training. Annotating includes syntactic and semantic annotations. Manning [13] defined the part-of-speech (POS) tagging as “the task of labeling (or tagging) each word in a sentence with its appropriate part of speech; we decide whether each word is a noun, verb, adjective, or whatever”. .Part-of-speech (POS) tagging is to assign one or more POS tags such as noun, verb, adjective, etc. to a lemma or synset (set of synonyms).

In this study, we consider the Arabic-English Al-Mawrid [14] dictionary as a parallel corpus of Arabic and English. The Al-Mawrid is not nearly annotated by any part-of-speech tags as stated by Fayed et al. [15, 16]. This study will exploit the translation equivalences (TEs) on the English side of the dictionary to assign part-of-speech tags to the Al-Mawrid senses. Assigning POS tags to senses of a bilingual lexicon is required in both HLT and NLP application. Furthermore, this step is required before translation of the English WordNet, linking a bilingual lexicon to the WordNet, or standardization of bilingual dictionary into WordNet-LMF.

The POS tagging task is composed of two steps. First, use the translation equivalences (TEs) of a sense and get the POS tags of them from the WordNet. Then, intersect the sets to acquire the most probable POS tags. The idea of this disambiguation process is that the most common POS tags among POS tags of TE are the most probable ones that represent the POS tags of a sense.

The contributions of this paper are:

- Proposing an independent-language algorithm that can be used to POS tag senses of any bilingual dictionary. This requires a repository of POS tags of the target language.
- Implementing and applying the algorithm on the Arabic-English Al-Mawrid lexicon.

## 2 Structures of Data Sources

### *A. Al-Mawrid*

The Arabic-English Al-Mawrid dictionary is a general-purpose dictionary. The headwords of the Al-Mawrid are arranged alphabetically according to the first letters. An entry of the Al-Mawrid starts by a bold headword. When a headword has more than one meaning or sense, each meaning occupies a subentry that is cited in separated lines. Subentries that contain collocations, idioms, terms, or examples are cited lately. A subentry consists of a mandatory Arabic section and an optional English section. The Arabic section consists of three fields that we name header, explanation, and cross-reference. The header is optional but the explanation and cross-reference are optional. A colon separates the header from its explanation. A dash may precede the cross-reference field. A subentry may express declaration, question, or exclamation [14, 15].

The three fields of an Arabic section have the same structure: each field consists of one or more words or phrases that are separated by either an Arabic comma or the conjunction word "أو". Comma-separated phrases are almost synonymous or near synonyms. The header has the headword of an entry if its subentry represents a sense of the headword. If a subentry does not represent a sense of the headword, it contains collocations, idioms, terms, or examples. The morphologic, syntactic, or semantic information is scattered in the header or explanation fields. The English section has one or more translation equivalence groups (TEGs) that are separated by semicolons. Each TEG has one or more translation equivalence (TE) phrases that are separated by comma. The phrases of a TEG are synonyms. The Al-Mawrid is not annotated by part-of-speech (POS) tags. Only very low number of two part-of-speech tags (approximately fifteen tags of nouns and adjectives) exists [14, 15]. Fig. 1 illustrates the microstructure of the Al-Mawrid.

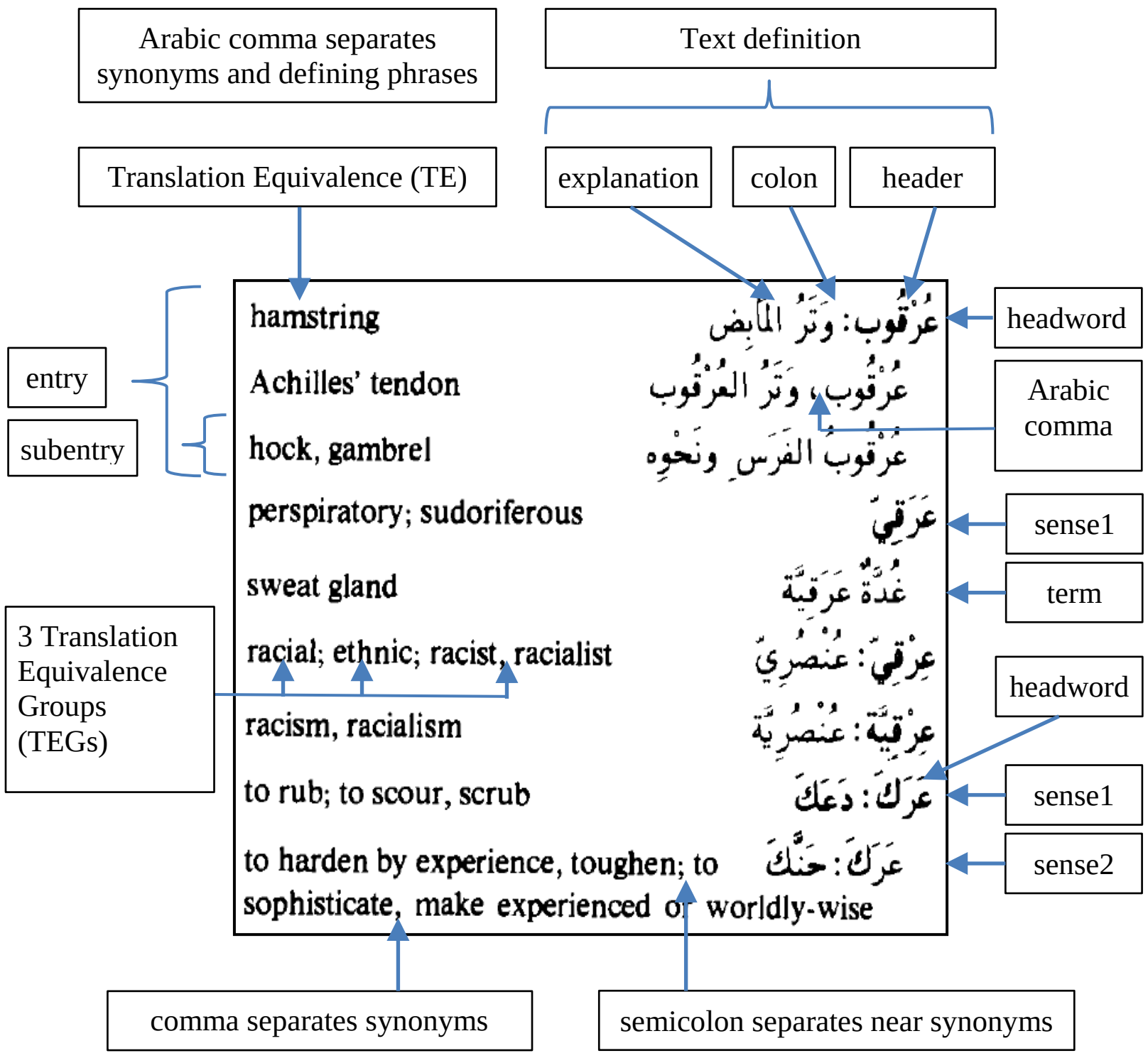

**Figure 1** The micro structure of the Al-Mawrid

### B. WordNet

WordNet [17] is a lexical system composed of an on-line English lexical database and software utilities. English concepts are organized into set of synonyms (synsets). Each synset is composed of words or phrases and is associated with glosses and illustrative examples. The database is divided into four categories: nouns, verbs, adjectives, and adverbs. Synsets are linked to other synsets by lexical and semantic relations. Fig. 2 and Fig. 3 show format of the WordNet sysnset [18] and examples respectively. Examples are excerpted from data.noun file and data.adj file of the WordNet database.

***synset_offset lex_filenum ss_type w_cnt word lex_id [word lex_id...] p_cnt [ptr...] [frames...] | gloss***

***synset_offset***
Current byte offset in the file represented as an 8 digit decimal integer.

***ss_type***
One character code indicating the synset type:
**n** NOUN
**v** VERB
**a** ADJECTIVE
**s** ADJECTIVE SATELLITE
**r** ADVERB

***word***
ASCII form of a word as entered in the synset by the lexicographer, with spaces replaced by underscore characters (_ ).

***gloss***
Each synset contains a gloss. A *gloss* is represented as a vertical bar (| ), followed by a text string that continues until the end of the line. The gloss may contain a definition, one or more example sentences, or both

**Figure 2** The format of the WordNet synset

**10502950** 18 **n** 02 **racist** 0 **racialist** 0 004 @ 09853645 n 0000 + 01155044 n 0202 + 06203758 n 0101 + 01155044 n 0101 | **a person with a prejudiced belief that one race is superior to others**

**01928283** 00 **s** 01 **racist** 0 001 & 01927654 a 0000 | **based on racial intolerance; "racist remarks"**

**00285905** 00 **s** 03 **racist** 0 **antiblack** 0 **anti-Semite**(a) 0 002 & 00285148 a 0000 + 09797742 n 0301 | **discriminatory especially on the basis of race or religion**

**Figure 3** Examples of synsets

The WordNet API search uses morphy function that preprocesses the searched string before looking up the database files. The preprocessing includes exceptional lists, morphological rules, collocations, hyphenations, etc. Table 1 contains suffixes that the morphy function uses to process the input string. For more explanation on the WordNet search function, see [18].

**TABLE -1**
RULES OF DETACHMENT

| POS | Suffix | Ending | POS | Suffix | Ending |
|---|---|---|---|---|---|
| NOUN | "s" | "" | VERB | "es" | "e" |
| NOUN | "ses" | "s" | VERB | "es" | "" |
| NOUN | "xes" | "x" | VERB | "ed" | "e" |
| NOUN | "zes" | "z" | VERB | "ed" | "" |
| NOUN | "ches” | "ch" | VERB | "ing" | "e" |
| NOUN | "shes” | "sh" | VERB | "ing" | "" |
| NOUN | "men” | "man" | ADJ | "er" | "" |
| NOUN | "ies" | "y" | ADJ | "est" | "" |
| VERB | "s" | "" | ADJ | "er" | "e" |
| VERB | "ies" | "y" | ADJ | "est" | "e" |

## 3 TAGGING ALGORITHM

Fig. 4 illustrates the general components of the proposed POS tagger. The main components are bilingual dictionary, monolingual dictionary or morphological analyzer of the second or target language, and POS disambiguator.

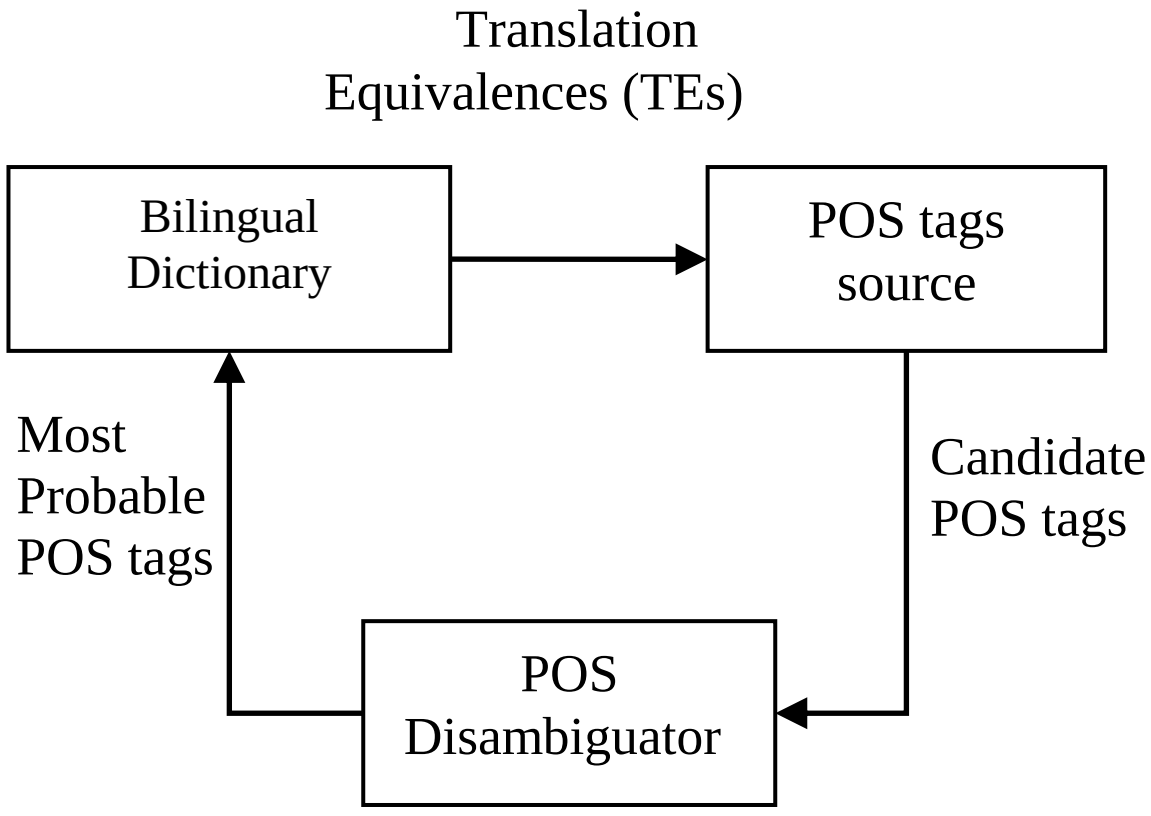

**Figure 4** Diagram of POS Tagging algorithm

Fig. 5 contains the pseudo code for the proposed POS tagging algorithm. Table 2 contains illustrative examples for the algorithm.

**For** each headword (HW) of a bilingual dictionary
    **For** each subentry sense (S) of HW
        **Get** the second language (SL) translation equivalences (TEs)
            **For** each TE word or phrase
                **Get** the part-of-speech (POS) tags set by consulting the SL dictionary or a SL morphological analyzer.
        **Intersect** the POS tag sets of all TEs
        **If** the result is not empty
            **Put** tags set equal to the result of intersection
        **Else**
            **Put** tags set equal to the most frequent POS tags

**Figure 5** Pseudo code Algorithm of Part-of-speech Tagging

We POS annotate senses of the Arabic-English Al-Mawrid dictionary by projecting the Tags from the English section to the Arabic section. The task accomplished by locking up WordNet database via the translation equivalence phrases, and then using a disambiguating method in the case of existing ambiguity in the POS tags. The disambiguation is simply accomplished by intersecting the sets of POS tags to get the common tag of sets. If the results of intersection are empty, the most frequent tag/tags will be the candidate POS tag of a sense. Table 2 contains illustrative examples for the algorithm.

**TABLE-2**
ILLUSTRATIVE EXAMPLE FOR ALGORITHM

| Mawrid | Intersection | WordNet |
|---|---|---|
| عَرَقِيّ<br>perspiratory<br>sudoriferous | $\{\emptyset\} \cap \{\emptyset\} = \emptyset$ | …<br>… |
| غُدَّةٌ عَرَقِيَّة<br>sweat gland | $\{n\}$ | (n) sweat gland, sudoriferous gland (any of the glands in the skin that secrete perspiration) |
| عِرْقِيّ: عُنْصُرِيّ<br>racial | $\{a, a\} \rightarrow \{2a\}$<br>{a} | (a) racial (of or related to genetically distinguished groups of people)<br>(a) racial (of or characteristic of race or races or arising from differences among groups) |
| عِرْقِيّ: عُنْصُرِيّ<br>ethnic | $\{n, a, a\} \rightarrow \{n, 2a\}$<br>{a} | (n) ethnic (a person who is a member of an ethnic group)<br>(a) cultural, ethnic, ethnical (denoting or deriving from or distinctive of the ways of living built up by a group of people)<br>(a) heathen, heathenish, pagan, ethnic (not acknowledging the God of Christianity and Judaism and Islam) |
| عِرْقِيّ: عُنْصُرِيّ<br>racist | $\{n, a, a\} \cap \{n\} = \{n\}$ | (n) racist, racialist (a person with a prejudiced belief that one racial group is superior to others)<br>(a) racist (based on racial intolerance)<br>(a) racist, antiblack, anti-Semite (discriminatory especially on the basis of race or religion) |
| عِرْقِيّ: عُنْصُرِيّ<br>racialist | | (n) racist, racialist (a person with a prejudiced belief that one racial group is superior to others) |
| عِرْقِيَّة: عُنْصُرِيَّة<br>racism | $\{n, n\} \cap \{n\} = \{n\}$ | (n) racism (the prejudice that members of one race are intrinsically superior to members of other races)<br>(n) racism, racialism, racial discrimination (discriminatory or abusive behavior towards members of another race) |
| عِرْقِيَّة: عُنْصُرِيَّة<br>racialism | | (n) racism, racialism, racial discrimination (discriminatory or abusive behavior towards members of another race) |

## 4 EXPERIMENTAL SETUP AND EVALUATION

### A. Dataset and Tools

We used the chapter Ayn "ع" of the Arabic-English Al-Mawrid dictionary [14] to evaluate the proposed algorithm that disambiguates part-of-speech tagging. The definitions of the Al-Mawrid are structured following the method of Diaa et al. [16]. In addition, we used the Princeton WordNet 3.0 [19, 20] as a source of part-of-speech tags of senses. We implemented the proposed algorithm in python and used the WordNet database that is implemented in Natural Language Toolkit (NLTK) [21].
In addition to preprocessing of the Al-Mawrid data in Diaa *et al*. [15, 16], we made additional preprocessing to the translation equivalences before used them in querying the WordNet API. Table 3 shows some of those modifications.

**TABLE -3**
EXAMPLES OF PREPROCESSING

| Sense | TE set |
|---|---|
| عَجَّلَ: حَثّ على العَجَلَة<br>to hurry, rush, urge, impel, press | { hurry, rush, urge, impel, press } |
| عَجَّجَ: أَثَارَ (الغُبَارَ)<br>to raise, swirl up (the dust) | {raise, swirl up} |
| العَجَلَةُ والجُزْع<br>wheel and axle | { wheel and axle, wheel-and-axle, wheel_and_axle, wheelandaxle } |

### B. Experiments

The WordNet API search functions make some morphological processing on the query word or phrase. We make some modifications on the WordNet interface and utilities codes. In each subsequent experiment, we augmented the steps by further modifications or adaptations on the previous experiment. Table 4 shows some modifications that can improve the accuracy of the algorithm.

**TABLE-4**
MODIFICATIONS TO IMPROVE THE ACCURACY OF THE ALGORITHM

| Experiment | Modification |
|---|---|
| 1 | • Suppress all the exception lookup.<br>• Suppress all the functions of morphology.<br>• Make the following preprocessing for the translation equivalences (TEs) phrases:<br>― remove "to " from beginning phrases that defining verbs,<br>― remove all parentheses,<br>― replace inner space in the collocations by score, underscore, space, and nothing. |
| 2 | Allow the morphology for plurals and apply rules in the morphology function. See **Table 1** for "Noun". |
| 3 | Set up the "Verb" tag as for any TE phrase starting by "to ". |
| 4 | Set up "Noun" as the default POS tag when the result of the algorithm is empty. The reason of that is the most frequent word class in any dictionary is the noun, making default POS tag will prevent empty results and increase coverage. |

### C. Evaluation Metrics and Evaluation Procedure

We used the precision, recall, and F-measure metrics [22, 23] to evaluate the proposed POS tagging algorithm. The definitions of the metrics are in equations (1), (2), and (3).

$$Precision = \frac{number\ of\ correct\ POS\ tags\ in\ tagged\ data}{number\ of\ correct\ POS\ tags\ in\ golden\ data} (1)$$

$$Recall = \frac{number\ of\ correct\ POS\ tags\ in\ tagged\ data}{number\ of\ total\ POS\ tags\ in\ tagged\ data} (2)$$

$$F = \frac{Precision\ \times Recall}{Precision\ + Recall} (3)$$

The evaluation procedure as following:

- Define a set of part-of-speech tags {noun, verb, adj, adv, phi} to define the senses of the dictionary. The first four senses are the POS tags of the WordNet. Phi is a POS tag for the undefined POS tags of senses. Examples of entries that take the Phi are sentences, verbal phrases, etc.
- first the senses of the Ayn chapter of the Al-Mawrid is tagged manually -as golden standards for evaluation,
- then the same chapter is tagged automatically following the proposed algorithm,
- finally, the *person* and *recall* are computed according to formulas.

## 5 RESULTS AND DISCUSSION

Table 5 contains the results of the four experiments. The first experiment is considered the base-line of the proposed algorithm. The values of precision and recall are moderate for the base-line experiment. The preci-sion and recall increased slightly by the experiment 2. However, in the experiment 3 and experiment 4, the values of precision and recall are increased dramatically.

**TABLE -5**
EVALUATION RESULTS

| Experiment | Precision | Recall | F |
|---|---|---|---|
| 1 | 71.58 | 74.67 | 36.55 |
| 2 | 72.30 | 75.18 | 36.86 |
| 3 | 89.12 | 87.84 | 44.24 |
| 4 | 93.10 | 87.36 | 45.07 |

The lesson of the previous results is that we can exploit the characteristics of the bilingual dictionary to increase the accuracy and coverage of the baseline algorithm.

## 6 RELATED WORKS

Our work is inspired by Pianta et al. [24] who developed an aligned multilingual database for the Italian language. They designed an assigning procedure that takes an input sense of an Italian word of the Italian-to-English section of the Collins dictionary and outputs a set of English candidate senses arranged by confidence scores. The confidence scores are computed based on a group of linking rule and each rule participates in the final score by weighted quantity. The Synset intersection is the linking rule that inspires our work. The synset intersection rule is based on the fact that TGRs may have multiple TEs which are synonymous. We can use other TEs to disambiguate an ambiguity of a TE. The rule takes different sets of candidates of TEs and intersects them. The candidates that are in the intersection get a partial confidence score.

Cucerzan and Yarowsky [25] bootstrapped a multilingual POS tagger using (1) an online or hard-copy pock-et-sized bilingual dictionary, (2) a basic library reference grammar, and (3) access to an existing monolingual text corpus in the language. As one step in the bootstrapping the POS tagger, they extracted a preliminary POS distributions from an untagged monolingual translation lists. For a given English translation word ei in the translation list (TL), the prior POS distribution probabilities are estimated from a large and balanced corpus. The combination of the Brown and WSJ corpora are used.

## 7 CONCLUSIONS AND FUTURE WORK

In this work, we proposed an independent-language algorithm that POS tags senses of a bilingual dictionary by using the translation equivalences of the target language and monolingual repository of senses. The algorithm is composed of two main steps: acquiring the sets of part-of-speech tags and then intersect them to get the POS tags that are common. We applied the algorithm on the Al-Mawrid Arabic-English dictionary and WordNet.

In future work, we will use more than one resource for part-of-speech tags and that will increase the accuracy and coverage. We plane also to link the senses of the A-Mawrid to the WordNet.

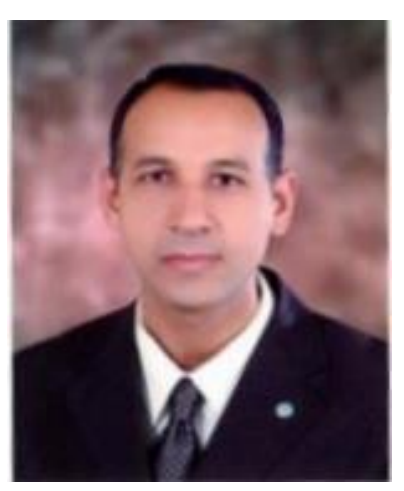

**Diaa El-Din Mohamed Abo-Fayed** received the B.E. degree in the Electronics, Faculty of Engineering, Mansoura University, 1995. He received M.Sc. degree in the Automatic Control from the Faculty of Engineering, Mansoura University, 2001, in the field of "Data Mining". Now Diaa is a PhD Candidate in the computer science in the Faculty of Computers and Information, Cairo University; the PhD is in the field of Arabic NLP. Diaa worked as a software research engineer in the COLTEC ME company in building HLT for Arabic. Diaa also worked in the News Group Co. as software research engineer. The News Group Co. specializes in the sourcing, distribution, creation, monitoring and analysis of news content in the emerging markets of the Middle East, Africa and the Indian sub-continent. Currently, Diaa is a communications engineer in the National Water Research Center.

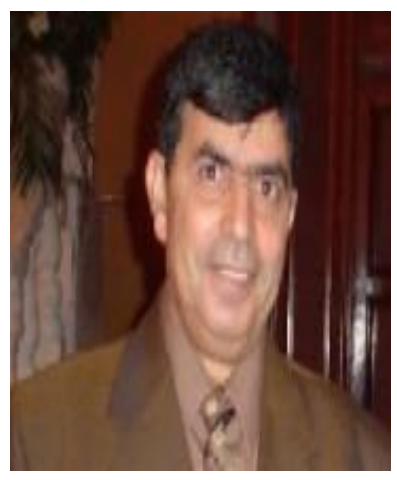

**Aly Aly Fahmy** is the former Dean of the Faculty of Computers and Information, Cairo University and a Professor of AI and ML, in the Department of Computer Science. He graduated from the Department of Computer Engineering, Technical College with honor degree. He specialized in Mathematical Logic. He received a master's degree from the ENSAE, Toulouse, France, 1976 in the field of Logical DB systems and then obtained his PhD from the Center for Studies and Research – CERT-DERI, 1979, Toulouse – France in the field of Artificial Intelligence. He participated in several national projects, built expert systems, and published many papers and books. Prof. Fahmy's main research areas are: Data and Text Mining, Mathematical Logic, Computational Linguistics, Text Understanding and Automatic Essay Scoring and Technologies of Man–Machine Interface in Arabic. He Directed the first Center of Excellence in Egypt in the field of Data Mining and Computer Modeling (DMCM). Prof. Aly Fahmy is currently involved in the implementation of the exploration project of Master's and Doctorate theses of Cairo University.

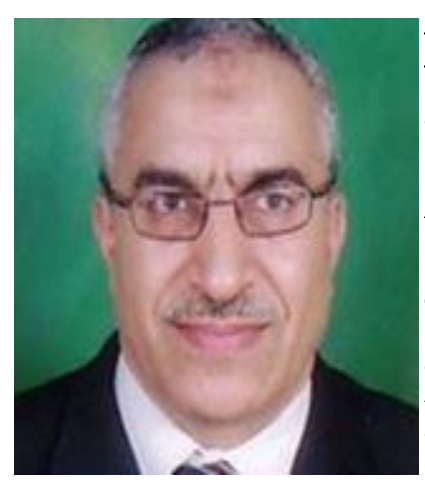

**Mohsen Abdelrazek Rashwan** received the B.Sc. and M.Sc. degrees in electronics and electrical communications from the Faculty of Engineering, Cairo University, Cairo, Egypt. The M.Sc. degree in systems and computer engineering from Carleton University, Ottawa, ON, Canada, and the Ph.D. degree in electronics and electrical communications from Queen's University, Kingston, ON, Canada. He currently serves as a Professor in the Department of Electronics and Electrical Communications, Faculty of Engineering, Cairo University, Cairo Egypt, and as the Managing Director of the Engineering Company for the development of Computer Systems that cofounded in 1993. Over the past research as well as building commercial products of Arabic NLP, digital speech processing, image processing, OCR, and e-learning. Among other several national and international mega projects, he has seved/been serving as a Senior Scientist in the EC's FP7 projects on Arabic HLT NEMLAR and MEDAR , and as a Co-PI in the Egyptian Data Mining and Computer Modeling Center of Excellence (DMCM-CoE).

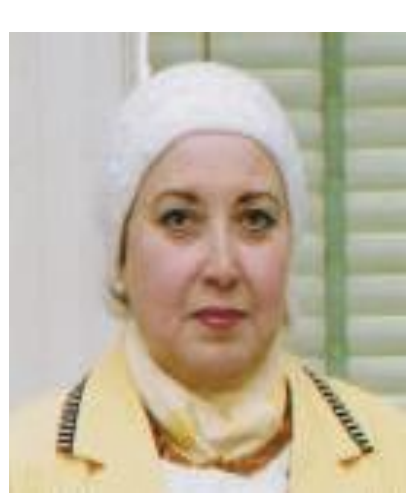

**Wafaa Kamel Fayed** B.A., M.A., Ph.D., Arabic Department, Faculty of Arts, Cairo University. Professor, Arabic Department - Faculty of Arts, Cairo University. Correspondent member, Arabic Language Academy, Damascus. Expert , Arabic Language Academy , Cairo. Supervisor of 49 Ph.D. and M.A. Degrees. Award of Ideal Professor at Cairo University. Honorary Certificate of distinction, 2004. Cairo University's Incentive award of human and social sciences, 2004. The 1st thesis using computer in applications of Arabic studies at 1974. Cairo University's appreciation award of humanities and pedagogic sciences 2013. Published 58 academic researchers at universal journals. The author of 7 academic books and translator of 2 books.

# توسيم معاني معجم عربي-إنجليزي بأنواع الكلمات بطريقة آلية

ضياء الدين محمد محمد أبوفايد*[1]، علي علي فهمي*[1]، محسن عبد الرازق رشوان**[2]، وفاء كامل فايد***[3]

كلية الحاسبات والمعلومات، قسم علوم الحاسب، جامعة القاهرة، الجيزة، مصر *

[1]diaa.fayed@outlook.com

[2]a.fahmy@fci-cu.edu.eg

كلية الهندسة، قسم الإلكترونيات والإتصالات الكهربية، جامعة القاهرة، الجيزة، مصر **

[3]*mrashwan@ieee.org*

كلية الآداب، قسم اللغة العربية وآدابها، جامعة القاهرة، الجيزة، مصر ***

[4]wafkamel@link.net

## ملخص

يقترح البحث خوارزمية لتوسيم معاني معجم عربي-انجليزي بأقسام الكلام بطريقة آلية. طبقت الخوارزمية على معجم المورد. تتم عملية التوسيم بنقل أقسام الكلام لمكافئات الترجمة الإنجليزية إلى معاني المعجم بعد عملية إزالة اللبس. يأخذ الخوارزم التوسيمات لمكافئات الترجمة من الوردنت. نحتاج توسيم معاني معجم عند ربط المعجم بالوردنت أو تحويل المعجم للصيغة القياسية WordNet-LMF حيث تكون مجموعة المعاني هي وحدة بناء المعجم وليس الكلمة.

بناء أدوات لعمل تطبيقات معالجة اللغات الطبيعية يتطلب خبراء واستثمارات ضخمة وفترات زمنية طويلة. فاذا كانت المقاربات المستخدمة احصائية فان ذلك يحتاج لذخائر لغوية ضخمة وموسمة؛ واذا كان المقاربات المستخدمة قاعدية، فان ذلك يحتاج لمعاجم ضخمة غنية بالمعلومات اللغوية. هذه المتطلبات المكلفة أدت لبزوغ مايسمى بالمقاربات المخفضة المصادر للغات الفقيرة في هذه المصادر لبناء أدوات تستخدم في تطبيقات معاجة اللغات الطبيعة.